\documentclass[preprint,12pt]{elsarticle}

\usepackage{amssymb}
\usepackage{amsmath}
\usepackage{amsfonts}

\usepackage{subcaption}
\usepackage{algorithmic}
\usepackage{algorithm}
\usepackage{tabularx,xcolor}
\usepackage{booktabs}   
\usepackage{multirow}  
\usepackage{hyperref}

\journal{Engineering Applications of Artificial Intelligence}

\begin{document}

\begin{frontmatter}



\title{\large Conditional Generative Framework with Peak-Aware Attention for Robust Chemical Detection under Interferences}


\author[1]{Namkyung Yoon\fnref{equal,orcidNY}}
\ead{nkyoon93@korea.ac.kr}

\author[1]{Sanghong Kim\fnref{equal,orcidSK}}
\ead{sanghongkim@korea.ac.kr}

\author[1]{Hwangnam Kim\corref{cor1}\fnref{orcidHK}}
\ead{hnkim@korea.ac.kr}

\fntext[equal]{These authors contributed equally to this work.}
\fntext[orcidNY]{ORCID: 0000-0002-2144-6664}
\fntext[orcidSK]{ORCID: 0009-0004-8643-8325}
\fntext[orcidHK]{ORCID: 0000-0003-4322-8518}
\cortext[cor1]{Corresponding author.}

\affiliation[1]{
    organization={School of Electrical Engineering, Korea University},
    city={Seoul},
    postcode={02841},
    country={Republic of Korea}
}

\begin{abstract}
Gas chromatography-mass spectrometry (GC-MS) is a widely used analytical method for chemical substance detection, but measurement reliability tends to deteriorate in the presence of interfering substances. In particular, interfering substances cause nonspecific peaks, residence time shifts, and increased background noise, resulting in reduced sensitivity and false alarms. To overcome these challenges, in this paper, we propose an artificial intelligence discrimination framework based on a peak-aware conditional generative model to improve the reliability of GC-MS measurements under interference conditions. The framework is learned with a novel peak-aware mechanism that highlights the characteristic peaks of GC-MS data, allowing it to generate important spectral features more faithfully. In addition, chemical and solvent information is encoded in a latent vector embedded with it, allowing a conditional generative adversarial neural network (CGAN) to generate a synthetic GC-MS signal consistent with the experimental conditions. This generates an experimental dataset that assumes indirect substance situations in chemical substance data, where acquisition is limited without conducting real experiments. These data are used for the learning of AI-based GC-MS discrimination models to help in accurate chemical substance discrimination.
We conduct various quantitative and qualitative evaluations of the generated simulated data to verify the validity of the proposed framework. We also verify how the generative model improves the performance of the AI discrimination framework. Representatively, the proposed method is shown to consistently achieve cosine similarity and Pearson correlation coefficient values above 0.9 while preserving peak number diversity and reducing false alarms in the discrimination model.
Thus, our proposed framework improves the robustness of GC-MS measurements and suggests a promising approach for reliable chemical detection in defense, environmental monitoring and industrial safety applications.
\end{abstract}

\begin{graphicalabstract}
\includegraphics[width=1.0\columnwidth]{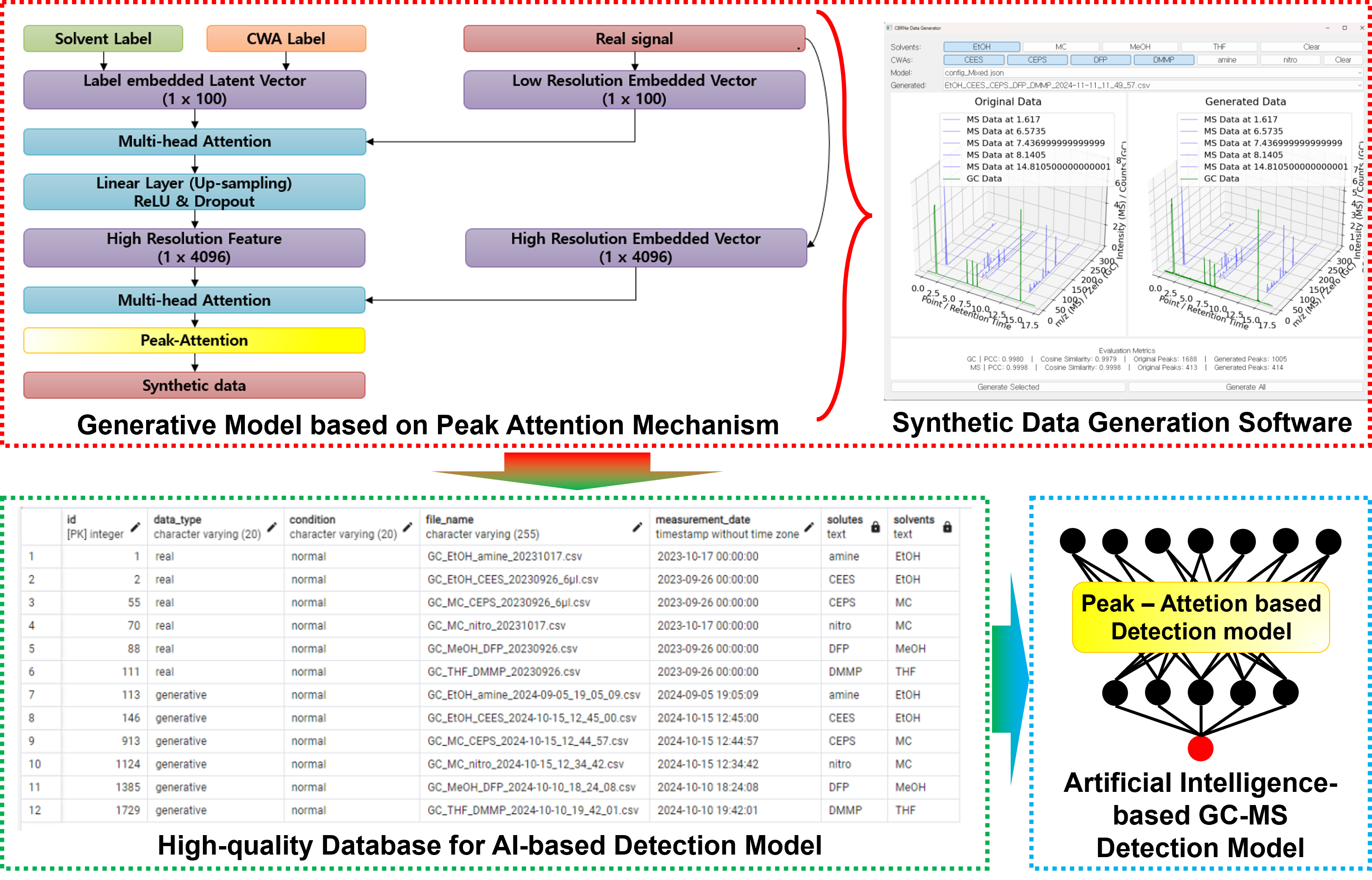}
\end{graphicalabstract}

\begin{highlights}
\item Developing a peak-aware generative framework with a novel peak-aware mechanism that selectively focuses on chemically meaningful regions in GC–MS and Raman spectra under diverse interference conditions.
\item Implementing an integrated environment that constructs large-scale synthetic spectral datasets and a unified SQL-based database for realistic chemical-agent detection experiments.
\item Verifying the proposed generative and contrastive-learning–based detection system through extensive evaluations, demonstrating improved robustness and accuracy across various chemical agents.
\end{highlights}


\end{frontmatter}




\section{Introduction}
Reliable detection and identification of complex hazardous chemicals in urgent situations is a critical requirement for appropriate initial responses in various environments, including defense and industry \cite{valdez2018analysis}.
Gas chromatography-mass spectrometry (GC-MS), which distinguishes chemicals based on their molecular mass, has long provided high sensitivity and selectivity as a standard for chemical analysis \cite{mavstovska2003practical}.
However, in realistic environments where substances interfere with fuel, solvents, soil, and building materials, such as real military or industrial conditions, the performance of GC-MS based detection systems is often degraded \cite{ranjan2023gas,eckenrode2001environmental}.

To ensure robustness against signal distortion against threshold setting and rule-based classification of traditional GC-MS spectra, the authors have taken advantage of these GC-MS data to conduct research on artificial intelligence (AI)-based detection algorithms to ensure higher identification performance \cite{yoon2024pioneering}.
Moreover, obtaining various GC-MS datasets is expensive and time-consuming, and is often limited by safety and security constraints. To address this issue, the authors have previously conducted artificial intelligence studies that analyze and apply variously on GC-MS data \cite{yoon2024dmgan, yoon2024unveiling}.

Against this background, this paper also introduces \textit{Peak-Attention Conditional Generative Framework}, which reflects the detailed peak structure of interfering materials and adapts to GC-MS data measured under interference conditions for more accurate identification.

The proposed method includes three main contributions:
\begin{itemize}
  \item \textbf{First.} We build an experimental GC-MS dataset of chemical substitute under various interference conditions, which progresses to exploratory data analysis to capture the key properties required by AI models.
  \item \textbf{Second.} We develop a novel peak-aware attention mechanism that selectively highlights local maximums in GC-MS data, allowing generative models to conditionally generate important spectral features that are often missed by existing models. Using this, we design a conditional generative adversarial network (CGAN) that encodes solvent and target chemical information into embedded latent vectors, enabling conditional data generation in line with the experimental configuration.
  \item \textbf{Third.} We gradually apply the generated synthetic GC-MS data to real data to train the discriminant model step by step, which verifies the performance improvement of the proposed discriminant framework.
\end{itemize}
We apply these frameworks to similar agents of chemical agents whose various measurements are restricted, taking into account real-world situations.
This study aims to advance the generation of GC–MS data for chemical agents across diverse environmental conditions and compound combinations, even in the absence of physical experiments. Furthermore, we investigate detection models that enhance the correspondence between synthetic and real-world data, thereby improving robustness and practical applicability in operational scenarios.

The following sections of the paper are organized as follows. Section 2 introduces the characteristics of GC-MS data, which are the principles based on the approach of the this study, various data generation techniques based on them, and the concepts that are the principles of the main deep learning techniques.
Section 3 describes the mechanism that focuses on the peak of the proposed GC-MS data and the framework structure that utilizes it. Section 4 introduces the measured data, the experimental environment, the experiments carried out under various conditions, and the results of the performance evaluation. Finally, Section 5 concludes the paper.

\section{Related Work}
\subsection{Limitation of GC-MS on Artificial Intelligence}
GC-MS is widely used as a reference technology for chemical identification due to its high sensitivity and performance. 
The components are separated and analyzed by GC-MS based on the mass characteristics of the mixed chemical molecules. The GC-MS measurement technique we use vaporizes the chemicals injected into the chromatography column, gets them on board the solvent, and records them as peaks in the chromatogram based on their mass. 
At this time, the chemical also interacts with the solvent, and our proposed framework reflects this as well\cite{mclafferty1978separation, hubschmann2025handbook, camarasu2000headspace}. 

Meanwhile, in the context of chemical detection, GC-MS is attracting attention with various AI technologies applied to analyze toxic industrial chemicals and chemical agents \cite{yoon2024pioneering}.
According to existing studies, GC-MS is a form of time series data, which tends to be limited in learning due to the large variation in the peak appearing in the retention time and the rest of the time of GC-MS \cite{yoon2024unveiling}.
We represented a tool to analyze the spectrum of chemicals measured through GC-MS developed in previous studies using the Matplotlib library, using it to represent the grafting results of the latest various time series algorithms \cite{yoon2024pioneering}:

\begin{figure*}[t]
  \centering
  \subfloat[Original GC-MS data\label{fig:original}]{
    \includegraphics[width=0.23\textwidth]{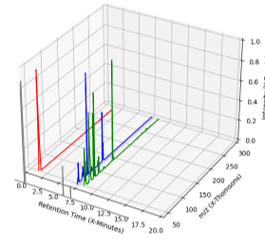}}
  \hfill
  \subfloat[TimeGAN\label{fig:timegan}]{
    \includegraphics[width=0.23\textwidth]{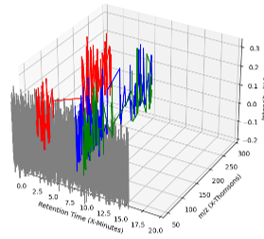}}
  \hfill
  \subfloat[LSTM-CNN GAN\label{fig:lstm-cnn-gan}]{
    \includegraphics[width=0.23\textwidth]{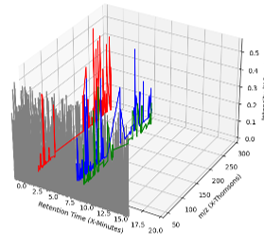}}
  \hfill
  \subfloat[DCGAN\label{fig:dCGAN}]{
    \includegraphics[width=0.23\textwidth]{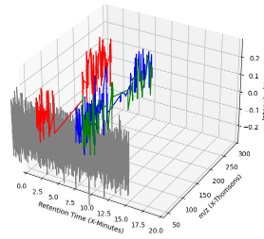}}
    \caption{Comparison of GC-MS spectral representations. 
    (a) Original experimental GC-MS data of 2-CEES with ethanol solvent. 
    (b)--(d) Limitations of artificial intelligence-based generative models based on GC-MS data characteristics (TimeGAN \cite{yoon2019time}, LSTM-CNN GAN \cite{zhu2019electrocardiogram}, DCGAN \cite{dewi2022synthetic}).}
    \label{gcms-comparison}
\end{figure*}

As shown in Fig. \ref{gcms-comparison}, we find that state-of-the-art AI models using conventional algorithms such as convolutional neural networks (CNNs) do not adequately reflect the variation in retention time and the characteristic peak patterns implied by mass-to-charge ratios (m/z) and learn with noise.
    
    

AI-generated models studied on common time series data show limitations in that it is difficult to learn the local maximum of the peak, as the difference in the size of the return time peak value and other peak values hinders the smooth learning of the AI \cite{yoon2024unveiling, yoon2024pioneering}.

Therefore, we propose an AI learning scheme that focuses on the peaks that appear at retention times in the proposed framework to overcome the limitations on the generation of synthetic data similar to real GC-MS data.
This can reliably obtain the desired GC-MS results without conducting actual experiments, which can be expected to improve the performance of the detection model and save human and time resources.

\begin{figure*}[]
\centering \makeatletter\IfFileExists{./figures/main.png}{\includegraphics[width=0.95\textwidth]{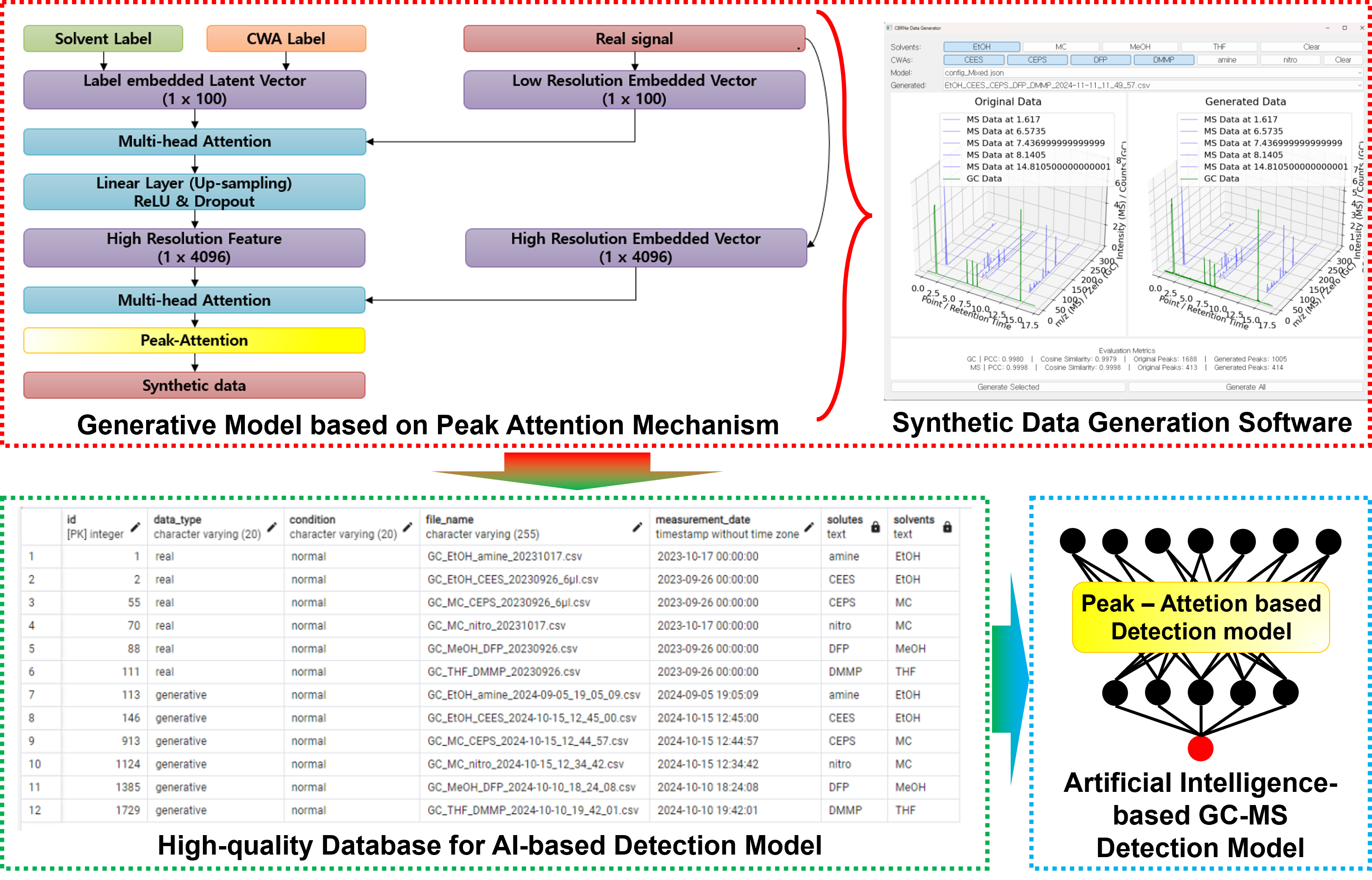}}{}
\makeatother 
\caption{{GC-MS Detection Model Framework Using peak-aware attention-Based Conditional Generation Model.}}
\label{main}
\end{figure*}

\subsection{AI-based measurements data approaches}
As briefly introduced in the previous section, recent advances in machine learning have led to various advances in AI-driven approaches for grafting to various measurement and instrumentation data, including GC-MS \cite{yoon2024dmgan}.
Data measured by various electronic devices have a one-dimensional form similar to time series data, and artificial intelligence technologies are being applied for such tasks as identification, outlier detection, and generation \cite{yoon2024detecting, yoon2024adaptive, yoon2023steam, yoon2025relaygan}.

Representatively, techniques such as autoencoders (AE) and variational autoencoders (VAEs) have been applied to eliminate signal noise and extract potential features \cite{berahmand2024autoencoders, singh2021overview}, and Generative Adversarial Networks (GANs) have demonstrated their ability to generate synthetic spectra similar to experimental data  \cite{audebert2018generative}. Nevertheless, even state-of-the-art high-performance generative models studied on data with the properties of typical time series data, as shown in Fig. \ref{gcms-comparison}, often fail to capture local peak properties that are important for chemical identification \cite{yoon2024pioneering}.

To overcome this, the main techniques underlying the proposed framework are newly designed attention mechanisms and data synthesis via conditional GANs (CGANs).
The attention mechanism is the underlying technique for the transformer models that make up recently widely used Large Language Mode (LLM) applications \cite{vig2019analyzing}.
Attention mechanisms are techniques at the heart of modern AI in deep learning that allow models to compute relevance between elements of input sequences by giving different weights to each query and key representation pair \cite{brauwers2021general}.
This allows the model to capture contextual dependencies more effectively than uniform weights. The attention distribution $\alpha_{ij}$ between a query $q_i$ and a key $k_j$ is given by:
\begin{equation}
\alpha_{ij} = \frac{\exp \left( \frac{q_i \cdot k_j}{\sqrt{d_k}} \right)}{\sum_{j'} \exp \left( \frac{q_i \cdot k_{j'}}{\sqrt{d_k}} \right)},
\end{equation}
where $d_k$ denotes the dimension of the key vector. The weighted sum of the values based on this score provides the output representation used in the subsequent layers inside the AI model.

The CGAN is an extension of a GAN in which two neural networks, generator and discriminator, are trained in a minimax game \cite{mirza2014conditional}. In traditional GANs, generator $G$ maps a random noise vector $z$ to synthetic data $G(z)$, while discriminator $D$ aims to distinguish the real sample $x$ from the generated sample. CGAN introduces the conditional information $c$ into both $G$ and $D$ to conditionally generate the data through the following optimization process:
\begin{multline}
\min_G \max_D \; \mathbb{E}_{x \sim p_{data}(x)}[\log D(x|c)] \\
+ \; \mathbb{E}_{z \sim p_z(z)}[\log (1 - D(G(z|c)|c))].
\end{multline}
This allows the generator of CGAN to generate samples for given conditions, improving both the diversity and controllability of the generated data to enhance the performance of the identification model.

In the following system design section, we propose an attention mechanism that focuses on the local peak properties of these time series data, which we describe as a framework that combines with CGAN to improve the performance of discriminative models.

\begin{figure}[t!]
\centering \makeatletter\IfFileExists{./figures/gc_ied.png}{\includegraphics[width=0.8\textwidth]{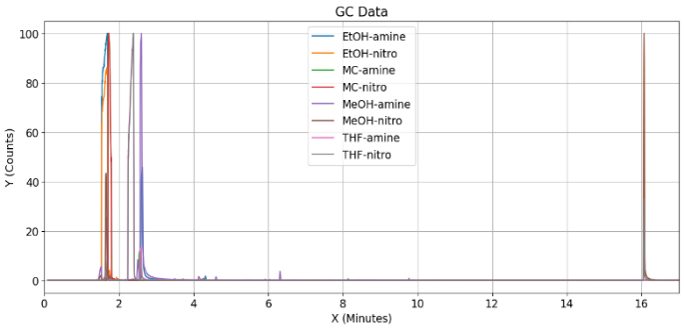}}{}
\makeatother 
\caption{{GC data of chemicals using four solvents.}}
\label{gc_ied}
\end{figure}
\begin{table}[t]
\centering
\caption{Analysis of GC data peak characteristics of chemical substances using four solvents.}
\label{gc-eda}
\resizebox{\columnwidth}{!}{%
\begin{tabular}{lccc}
\hline
\textbf{Condition}  & \textbf{Total peak area} & \textbf{Mean intensity} & \textbf{Standard deviation} \\
\hline
Ethylenediamine\_EtOH    & 15.54 & 1.16 & 9.26 \\
4-Nitrophenol\_EtOH     & 19.11 & 0.97 & 8.29 \\
Ethylenediamine\_MC      & 11.48 & 0.52 & 6.10 \\
4-Nitrophenol\_MC    & 13.46 & 0.50 & 6.08 \\
Ethylenediamine\_MeOH    & 14.48 & 0.45 & 5.02 \\
4-Nitrophenol\_MeOH    & 14.65 & 0.27 & 3.54 \\
Ethylenediamine\_THF     & 11.89 & 0.28 & 7.67 \\
4-Nitrophenol\_THF     & 13.88 & 0.81 & 7.64 \\
\hline
\end{tabular}%
}
\end{table}

\section{System Design}

The GC-MS detection model framework using the generative model based on peak-aware attention proposed in this paper is designed as shown in Fig. \ref{main} to improve the measurement reliability of GC-MS data detection model.
The entire framework consists of three main components: (i) the newly proposed peak-aware attention mechanism for GC-MS data characteristics, (ii) the conditional generative model with the peak-aware attention mechanism, and (iii) the detection model with the database built through the generative model and the peak-aware attention.
The framework aims to integrate instrumentation knowledge of GC-MS peak properties with advanced generative modeling techniques to increase the fidelity of synthetic signal generation and save various resources while reducing false alarms in chemical detection tasks.

\subsection{Peak-aware attention mechanism}
Recently, attention mechanisms are actively used in most fields as a technique to improve learning performance by emphasizing important information in the data. However, since it is difficult to secure sufficient performance in GC-MS data as shown in Section 2 simply by applying the existing attention mechanism, we propose a new attention mechanism optimized for the characteristics of the GC-MS data.

Fig. \ref{gc_ied} is the result of the measurement of four solvents - Ethanol (EtOH), Methanol (MeOH), Methylene chloride (MC), Tetrahydrofuran (THF)- used in experiments and verification in this study by combining dangerous chemicals (ethylenediamine and 4-nitrophenol), and only GC appears as a two-dimensional picture for peak visibility.
At this time, it is difficult to visually identify the two chemicals as the solvent changes, but as shown in Table \ref{gc-eda}, the characteristics such as the number of peaks, the total peak area and the standard deviation of the peak intensity are all different.

Therefore, in order to respond to the characteristics of GC-MS that require different detection according to slightly different peaks, the peak-aware attention mechanism selectively emphasizes importance based on the rate of change in the value of the GC-MS peak.

The proposed peak-aware attention mechanism  is designed to emphasize regions of the GC–MS spectrum exhibiting significant signal variations, which typically correspond to chemically informative peaks. Concretely, given a one-dimensional signal $x = [x_1, x_2, \dots, x_T]$ of length $T$, the slope of the local peak is calculated as the absolute difference between adjacent points:
\begin{equation}
s_t = |x_{t} - x_{t-1}|, \quad t = 2, \dots, T.
\end{equation}
These slopes are then exponentially scaled to magnify sharp changes, followed by a normalization step across the entire sequence to obtain the attention weights:
\begin{equation}
\alpha_t = \frac{\exp(s_t)}{\sum_{j=2}^{T} \exp(s_j)}, \quad \sum_{t=2}^{T} \alpha_t = 1.
\end{equation}
For alignment with the original signal length, a zero-padding term is appended, and the weights are further refined through a learnable $1$-D convolution layer with a sigmoid activation:
\begin{equation}
\tilde{\alpha} = \sigma(\text{Conv1D}(\alpha)),
\end{equation}
where $\sigma(\cdot)$ denotes the sigmoid nonlinearity. The resulting attention weights $\tilde{\alpha}$ are reshaped to match the dimension of $x$, enabling element-wise reweighting of the GC-MS signal.

This formulation effectively suppresses low-variation background noise while adaptively enhancing high-slope regions corresponding to peak structures. By embedding the slope of peak-based exponential weighting into a trainable convolutional mapping, the mechanism combines deterministic peak detection with learnable feature refinement. Consequently, the generator can focus on reproducing fine-grained peak variations that conventional attention or GAN architectures often overlook, thereby improving the fidelity of synthetic GC-MS data generation.

This mechanism suppresses background noise while giving higher weight to the informed peaks. With this design, the generator aims to reproduce fine-grained peak features that traditional GAN architectures often fail to capture.
\begin{algorithm}[]
\caption{Conditional GC-MS Generative and Detection Framework with peak-aware attention mechanism}
\label{peak_attention_CGAN}
\small
\begin{algorithmic}[1]
\REQUIRE Real GC-MS spectra $x$, solvent label $c_s$, solute label $c_t$, noise vector $z$
\ENSURE Synthetic GC-MS spectra $\hat{x}$, trained detection model $M$

\STATE \textbf{Initialization:}
Initialize generator $G$, discriminator $D$, peak-aware attention weights $\tilde{\alpha}$, and detection model $M$.
Set learning rates $\eta_G, \eta_D$ and trade-off hyperparameters $\lambda, \mu$.

\STATE \textbf{Embedding:}
Map condition labels $(c_s,c_t)$ into latent embeddings $E_c \in \mathbb{R}^{d}$ via compositional encoding.

\STATE \textbf{Generator Forward Pass:}
\STATE \quad 1) Fuse embeddings with multi-head attention: $H_1 = \mathrm{MHA}(E_c)$
\STATE \quad 2) Concatenate $H_1$ with noise vector $z$
\STATE \quad 3) Upsample via linear projection and ReLU/dropout to obtain $F_{\mathrm{up}} \in \mathbb{R}^{T \times d}$
\STATE \quad 4) Refine with second multi-head attention: $H_2 = \mathrm{MHA}(F_{\mathrm{up}})$
\STATE \quad 5) Apply peak-aware attention weighting: $\hat{X} = H_2 \odot \tilde{\alpha}$
\STATE \quad 6) Generate synthetic spectra $\hat{x} = G(\hat{X})$

\STATE \textbf{Discriminator Forward Pass:}
\STATE \quad 1) Compute $D(x,c)$ for real spectra and $D(\hat{x},c)$ for generated spectra
\STATE \quad 2) Extract multi-scale features with convolution + attention

\STATE \textbf{Loss Functions:}
\STATE \quad $\displaystyle
\mathcal{L}_{G} =
\mathbb{E}_{z,c}\!\left[ -\log D(G(z\mid c),c) \right]
+ \lambda \,\mathbb{E}_{x,\hat{x}\mid c}\!\left[
\bigl\|\mathrm{STFT}(x)-\mathrm{STFT}(\hat{x})\bigr\|_2^2
\right]$
\STATE \quad $\displaystyle
\mathcal{L}_{D} =
\frac{1}{2}\,\mathbb{E}_{x,c}\!\left[(D(x,c)-1)^2\right]
+ \frac{1}{2}\,\mathbb{E}_{z,c}\!\left[D(G(z\mid c),c)^2\right]$

\STATE \textbf{Optimization:}
Update $G \leftarrow G - \eta_G \nabla_G \mathcal{L}_G$,
$D \leftarrow D - \eta_D \nabla_D \mathcal{L}_D$.

\STATE \textbf{Database Integration:}
Store $\{\hat{x},c_s,c_t,\text{timestamp}\}$ in an SQL-based schema
with fields \texttt{(id, data\_type, condition, solvent, solute, date, file\_name)}
to enable query and retrieval of real and synthetic spectra.

\STATE \textbf{Detection Model Training:}
\STATE \quad 1) Extract peak-aware features
$f(x)=\sum_{t}\tilde{\alpha}_t h_t$
\STATE \quad 2) Compute posterior
$p(y\mid x)=\mathrm{softmax}(W f(x)+b)$
\STATE \quad 3) Train detection model $M$ on combined real and synthetic data.

\RETURN Synthetic data $\hat{x}$ and trained detection model $M$
\end{algorithmic}
\normalsize
\end{algorithm}

\subsection{Conditional GC-MS generative model}

The GC-MS data generation framework is formulated as a CGAN, where solvent and target chemical labels are transformed into embedded latent vectors through compositional encoding. These latent vectors provide numerical representations of experimental conditions, enabling domain-aware GC-MS data generation. The conditioned embedded latent vector is linked to random noise to produce synthetic GC-MS data that matches the label specified by CGAN.

To effectively capture dependencies between condition embeddings and GC-MS data features, the generator adopts a dual multi-head attention mechanism \cite{cordonnier2020multi}. In the first stage of the generative model neural network shown in Fig. \ref{main}, solvent and target embeddings are fused into a joint latent representation:
\begin{equation}
H_1 = \text{MHA}(Q=E_c, K=E_c, V=E_c),
\end{equation}
where $E_c \in \mathbb{R}^{d \times h}$ denotes the concatenated conditional embeddings. In the second stage, after up-sampling the latent vectors into high-resolution features \cite{quan2024deep}, a second multi-head attention module refines long-range dependencies:
\begin{equation}
H_2 = \text{MHA}(Q=F_{up}, K=F_{up}, V=F_{up}),
\end{equation}
with $F_{up} \in \mathbb{R}^{T \times d}$ representing the up-sampled feature matrix.

Furthermore, to ensure that chemically informative peaks are not suppressed, the proposed peak-aware attention mechanism weights the refined features with scores $\tilde{\alpha}$ of peak-aware attention mechanism:
\begin{equation}
\hat{X} = H_2 \odot \tilde{\alpha},
\end{equation}
where $\odot$ denotes element-wise multiplication. These selective weights, based on the peak-aware attention mechanism, highlight local peak structures that are important for chemical discrimination while suppressing background noise.

The CGAN constructed in this way is adversarially trained \cite{smith2020conditional}, where the generator learns to generate GC-MS data by focusing on realistic peaks, and the discriminator distinguishes between real and synthetic data.
In this case, the generator is optimized with a composite objective that integrates binary cross-entropy (BCE) and spectral reconstruction via the short-time Fourier transform (STFT) as follows \cite{haber2017simple, durak2003short, hodson2022root}:
\begin{multline}
\mathcal{L}_{G} = 
\mathbb{E}_{z \sim p_z(z), \, c \sim p(c)} 
\Big[ - \log D(G(z|c), c) \Big] \\
+ \lambda \, \mathbb{E}_{x \sim p_{data}(x), \, \hat{x} \sim p_G(x|c)} 
\Big[ \| \text{STFT}(x) - \text{STFT}(\hat{x}) \|_2^2 \Big].
\end{multline}

In addition, the discriminator is optimized using the least-squares adversarial formulation, where the prediction error is measured by the mean squared error (MSE) \cite{hodson2022root}. 

\begin{multline}
\mathcal{L}_{D} =
\frac{1}{2} \, \mathbb{E}_{x \sim p_{data}(x), \, c \sim p(c)} 
\Big[ (D(x,c) - 1 )^2 \Big] \\
+ \frac{1}{2} \, \mathbb{E}_{z \sim p_z(z), \, c \sim p(c)} 
\Big[ (D(G(z|c),c))^2 \Big].
\end{multline}

Through the integration of synthetic embeddings, double-head attention, peak-aware attention, and composite adversarial-spectrum loss, the proposed CGAN conditionally generates realistic but hypothetical synthetic GC-MS data while maintaining peak diversity.

\subsection{System workflow with detection model}
As shown in Fig. \ref{main}, we propose for the discriminant framework, the peak-aware attention mechanism-based CGAN generation model builds SQL-based AI-based software that learns and generates on real-world measured data.
All data synthesized through generative software and related metadata such as state, solvent, solute, and timestamps are stored in relational databases to enable efficient query, search, and large-scale evaluation. With this structured repository, in the detection phase, we train for the optimization of peak-aware attention based classifiers on both real and synthetic GC-MS data.
The structure of the peak-aware attention based classifier treats GC data and MS data as independent streams, respectively, and in both cases, CNN-based preprocessing and transformer encoders are combined to simultaneously learn local patterns and long-term dependencies.

In the GC data stream, a peak candidate of GC data about the time axis is calculated via a one-dimensional CNN layer consisting of seven kernels, three padding, and a 128-dimensional transformer encoder consisting of four heads and two layers. Then, it outputs the peak presence probability through a fully connected (FC) layer and learns it as a BCE loss.
Furthermore, we extract the mass-charge ratio ($m/z$) feature with two one-dimensional CNN layers consisting of seven and five kernels, respectively, for the MS data stream. Then, it passes through a 128-dimensional transformer encoder consisting of an FC layer, four heads, and two layers to output a class of chemicals from the final FC head and learn with a multi-label BCE loss.

At this time, given the input spectrum $x$ for both GC and MS data streams respectively, the model computes the peak-aware feature representation $f(x)$ by applying the attention weight $\tilde{\alpha}$:
\begin{equation}
f(x) = \sum_{t=1}^T \tilde{\alpha}_t \, h_t, \quad \sum_{t=1}^T \tilde{\alpha}_t = 1,
\end{equation}
where $h_t$ denotes hidden features at retention index $t$. This weighted aggregation ensures that discriminative peaks dominate the learned representation.

In addition, the output of the softmax-based detector at the FC head used in each stream is as follows:
\begin{equation}
p(y|x) = \text{softmax}(W f(x) + b),
\end{equation}
where $W$ and $b$ are trainable parameters.

In the following experimental sections, the performance of the overall detection framework, including the proposed generative model, is verified using various metrics such as cosine similarity, Pearson correlation coefficient, peak count matching, and accuracy.

\section{Experiments}


\begin{table}[]
\centering
\caption{Network and Training Settings}
\label{settings}
\begin{tabular}{p{0.4\columnwidth}|p{0.5\columnwidth}}
\hline
\textbf{Network Parameters} & \\
\hline
Solvent label dimension & 4 \\
Solute label dimension  & 6 \\
Embedding dimension     & 100 \\
Final output dimension  & 5347 \\
Generator depth         & 16 \\
Hidden dimension        & 32 \\
\hline
\textbf{Training Parameters} & \\
\hline
Iterations              & 100,000 \\
Generator learning rate & $1 \times 10^{-4}$ \\
Discriminator learning rate & $1 \times 10^{-5}$ \\
Batch size              & 128 \\
\hline
\end{tabular}
\end{table}
We implemented the framework proposed in Section 3 using Python code to implement neural networks as shown in Table~\ref{settings} in the intel i9-12900K, 64GB RAM, and RTX 3090 (24GB) environment.
Through this, we aim to build a database where the proposed framework can contribute to different fields and verify that stable detection is possible even in interfering materials or complex mixed situations.

\subsection{Dataset description}

We constructed a GC-MS database containing chemicals and interfering substances to evaluate and measure the proposed framework under realistic conditions.
The database covers several categories of chemical warfare surrogates, explosive-related substances, and solvents commonly encountered in industrial and military field environments.
At this time, the actual acquisition of the nerve agent and the blister agent is limited by the system, so they are measured using a molecular structure similar to the agent.
Based on Table~\ref{gcms_targets}, the chemicals used in the experiment can be divided into categories of blisters, nerve agents, and improvised explosive devices (IED) as follows:

\begin{itemize}
    \item Nerve agents are highly toxic organophosphate compounds in the nervous system that can often be used as chemical weapons, such as Sarin, VX gas. To counter the situation of Sarin, VX gas, a colorless, odorless chemical that is difficult to discriminate immediately, we used Dimethyl Methylphosphonate (DMMP) and Diisopropyl fluorophosphate (DFP), which are legally obtainable but have nearly similar molecular structures~\cite{zheng2010advances, wymore2014hydrolysis}.
    \item Blister agents are toxic chemicals that cause burns-like blisters with severe pain when exposed to the skin. We used 2-chloroethyl sulfides (2-CEES) and 2-chloroethyl phenyl sulfides (2-CEPS), whose molecular structures are nearly similar, to detect accidents occurring primarily through the Mustard gas agents group in industry or wartime sulfuric acid~\cite{kim2021decomposition, dubey2009thiocompounds}.
    \item IED is an Improvised Explosive Device (IED), which means improvised explosives or homemade bombs, and we used 4-Nitrophenol, which in the past pollutes the soil near explosives, textile factories, and military facilities, and ethylenediamine, which can be exploited as colorless and powerful explosives~\cite{apak2020colorimetric, choodum2017selective}.
\end{itemize}

\begin{table}[t]
\centering
\caption{GC-MS target substances and interfering agents}
\label{gcms_targets}
\resizebox{\columnwidth}{!}{%
\begin{tabular}{l l l}
\hline
\textbf{Category} & \textbf{Examples} & \textbf{GC-MS Target Compounds} \\
\hline
Nerve agent & Sarin, VX & DMMP, DFP \\
Blister agent & Mustard gas & 2-CEES, 2-CFPS \\
IED & -- & 4-Nitrophenol, Ethylenediamine \\
\hline
\end{tabular}%
}
\end{table}
\begin{table}[]
\centering
\caption{GC-MS experimental conditions under interfering environments}
\label{gcms_conditions}
\begin{tabular}{l p{0.5\columnwidth}}
\hline
\textbf{Condition} & \textbf{Description} \\
\hline
Solvent & Ethanol (EtOH), Methanol (MeOH), Methylene chloride (MC), Tetrahydrofuran (THF) \\
Reaction setup & Solvent + interfering material reacted 24 hours \\
Interfering materials & Brick, soil, grass, asphalt, kerosene, acetone \\
Excluded high-risk cases & Combinations of IED-related materials with brick, cement, acetone, gasoline \\
\hline
\end{tabular}
\end{table}

Experimental settings based on interference conditions for GC-MS detection in various emergency scenarios have been set up as follows:

\begin{itemize}
    \item The concentration of all interfering substances was varied according to Table~\ref{gcms_conditions} and combined with toxic chemical surrogates for measurement.
    \item Brick, soil, grass, and asphalt were mixed with hydrofluoric acid (5 ml), reacted for 24 hours, filtered, and used in subsequent GC-MS measurements.
    \item Kerosene and acetone were used as contact solvents, sampled after 24 hours of reaction.
    \item For safety reasons, interfering substances were excluded from interference experiments because IEDs such as 4-nitrophenol, ethylenediamine could cause explosions, toxic gases, or uncontrolled reactions.
\end{itemize}

\begin{table}[htbp]
\centering
\tiny                           
\setlength{\tabcolsep}{2.5pt}        
\renewcommand{\arraystretch}{0.9}     
\caption{Integrated evaluation of generated GC-MS data. Global similarity (GC evaluation) and local MS evaluation at representative retention times ($t$).}
\label{integrated_quality}
\begin{subtable}{\linewidth}
\centering
\caption{GC Evaluation}
\begin{tabularx}{\linewidth}{lccc}
\toprule
\textbf{Condition} & \textbf{PCC} & \textbf{Cosine} & \textbf{Peaks (Real/Gen)} \\
\midrule
4-nitrophenol + EtOH  & 0.99 & 0.99 & 2/2 \\
4-nitrophenol + MC    & 0.99 & 0.99 & 3/2 \\
4-nitrophenol + MeOH  & 0.99 & 0.99 & 2/2 \\
4-nitrophenol + THF   & 0.99 & 0.99 & 2/2 \\
Ethylenediamine + EtOH & 0.99 & 0.99 & 2/2 \\
Ethylenediamine + MC   & 0.99 & 0.99 & 3/3 \\
Ethylenediamine + MeOH & 0.99 & 0.99 & 4/3 \\
Ethylenediamine + THF  & 0.99 & 0.99 & 2/2 \\
2-CEES + EtOH          & 0.94 & 0.94 & 2/2 \\
2-CEES + MC            & 0.99 & 0.99 & 2/2 \\
2-CEES + MeOH          & 0.97 & 0.96 & 4/6 \\
2-CEES + THF           & 0.98 & 0.98 & 2/2 \\
2-CEES + 2-CEPS + EtOH & 0.99 & 0.99 & 3/3 \\
2-CEES + 2-CEPS + DFP + MeOH & 0.99 & 0.99 & 5/5 \\
2-CEPS + DFP + DMMP + THF    & 0.99 & 0.99 & 5/4 \\
2-CEES + 2-CEPS + DFP + DMMP + MC & 0.99 & 0.99 & 6/6 \\
\bottomrule
\end{tabularx}
\end{subtable}

\vspace{0.2em}   

\begin{subtable}{\linewidth}
\centering
\caption{MS evaluation at representative retention times}
\renewcommand{\arraystretch}{0.85}    
\begin{tabularx}{\linewidth}{lcccccccc}
\toprule
\textbf{Condition}
& \textbf{$t_1$ (min)} & \textbf{PCC} & \textbf{Cosine} & \textbf{Peaks}
& \textbf{$t_2$ (min)} & \textbf{PCC} & \textbf{Cosine} & \textbf{Peaks} \\
\midrule
4-nitrophenol + EtOH  & 16.07 & 1.00 & 1.00 & 8/8   & --    & --   & --   & --   \\
4-nitrophenol + MC    & 16.07 & 0.99 & 0.99 & 7/7   & --    & --   & --   & --   \\
4-nitrophenol + MeOH  & 16.07 & 1.00 & 1.00 & 7/8   & --    & --   & --   & --   \\
4-nitrophenol + THF   & 16.07 & 1.00 & 1.00 & 8/8   & --    & --   & --   & --   \\
Ethylenediamine + EtOH &  2.63 & 0.99 & 0.99 & 2/2   & --    & --   & --   & --   \\
Ethylenediamine + MC   &  2.57 & 1.00 & 1.00 & 2/2   & --    & --   & --   & --   \\
Ethylenediamine + MeOH &  2.60 & 1.00 & 1.00 & 2/2   & --    & --   & --   & --   \\
Ethylenediamine + THF  &  2.59 & 1.00 & 1.00 & 2/2   & --    & --   & --   & --   \\
2-CEES + EtOH          &  1.60 & 1.00 & 1.00 & 2/2   & 7.44  & 1.00 & 1.00 & 5/5 \\
2-CEES + MC            &  1.74 & 1.00 & 1.00 & 7/7   & 7.43  & 1.00 & 1.00 & 5/5 \\
2-CEES + MeOH          &  1.50 & 0.99 & 0.99 & 13/13 & 7.18  & 1.00 & 1.00 & 6/6 \\
2-CEES + THF           &  2.31 & 0.99 & 0.99 & 3/3   & 7.43  & 1.00 & 1.00 & 5/5 \\
2-CEES + 2-CEPS + EtOH &  1.61 & 1.00 & 1.00 & 2/2   & 14.81 & 0.99 & 0.99 & 10/10 \\
2-CEES + 2-CEPS + DFP + MeOH & 1.65 & 0.99 & 0.99 & 2/2   & 7.43  & 1.00 & 1.00 & 5/5 \\
2-CEPS + DFP + DMMP + THF    & 6.56 & 0.99 & 0.99 & 7/7   & 8.13  & 1.00 & 1.00 & 2/2 \\
2-CEES + 2-CEPS + DFP + DMMP + MC & 6.56 & 1.00 & 1.00 & 7/7 & 7.43 & 1.00 & 1.00 & 5/5 \\
\bottomrule
\end{tabularx}
\end{subtable}

\end{table}

\begin{figure*}[]
\centering
\subfloat[Real Single Data]{%
    \includegraphics[width=0.25\textwidth]{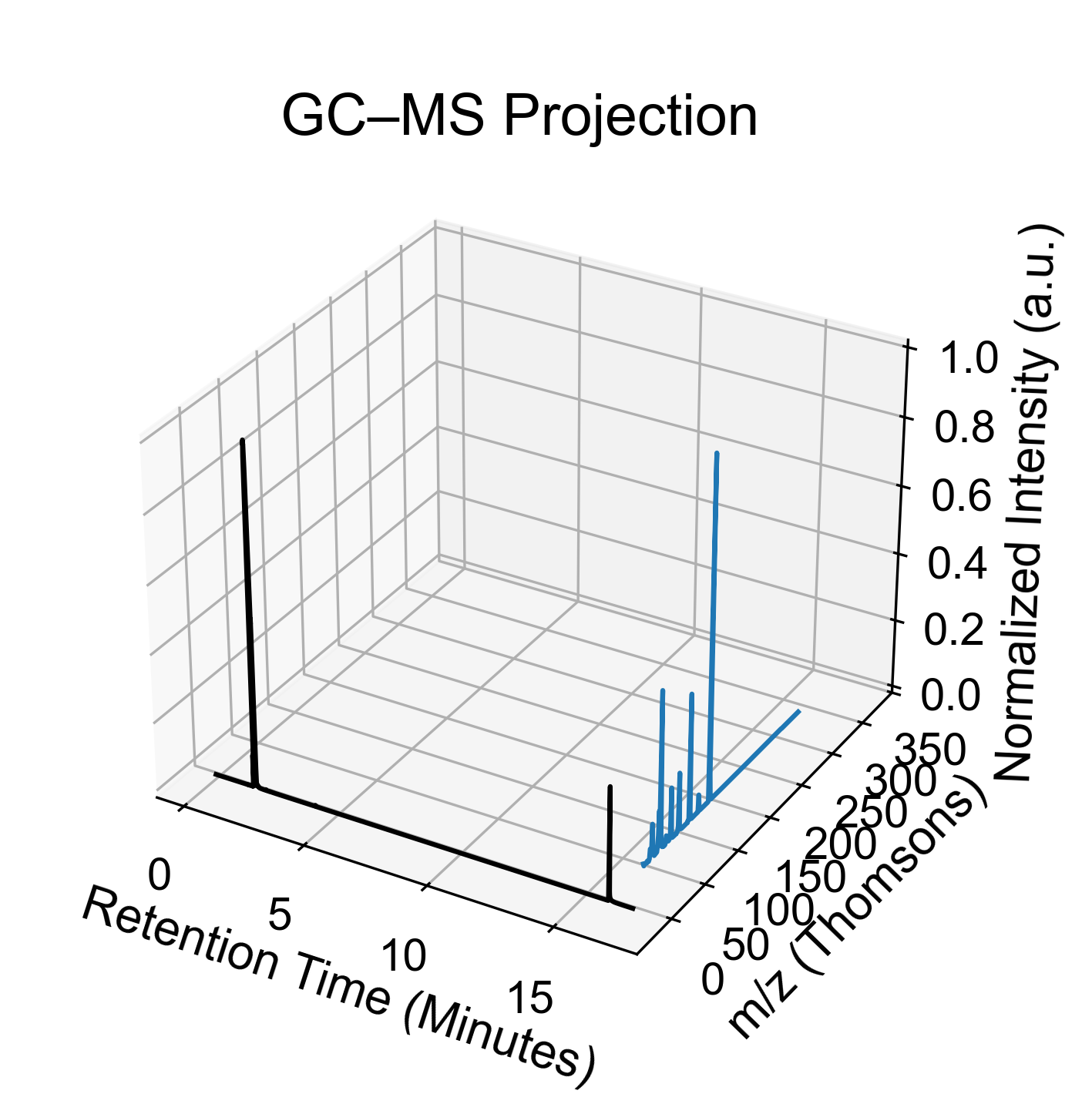}}
\subfloat[Generated Single Data]{%
    \includegraphics[width=0.25\textwidth]{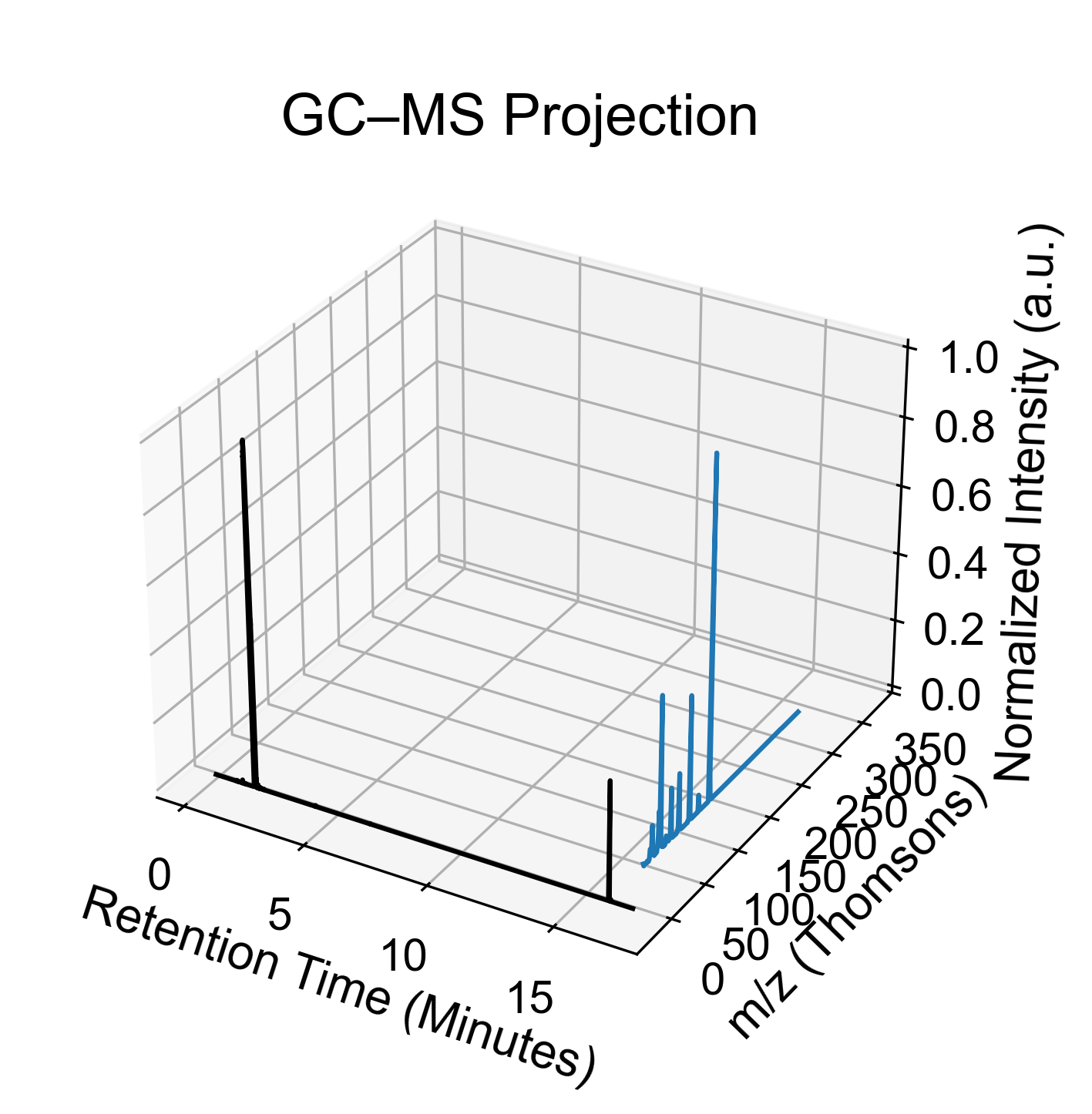}}
\subfloat[Real Mixed Data]{%
    \includegraphics[width=0.25\textwidth]{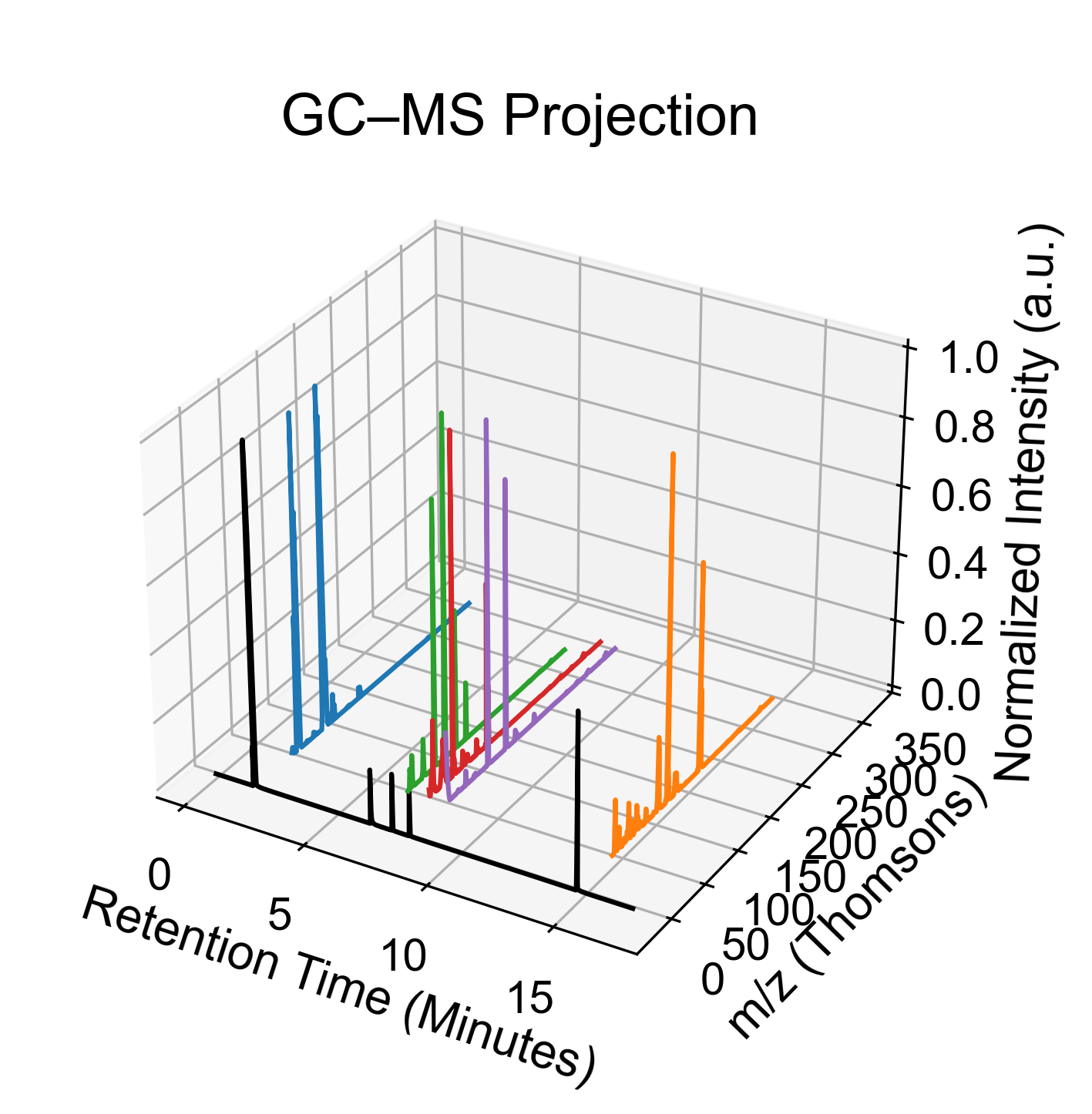}}
\subfloat[Generated Mixed Data]{%
    \includegraphics[width=0.25\textwidth]{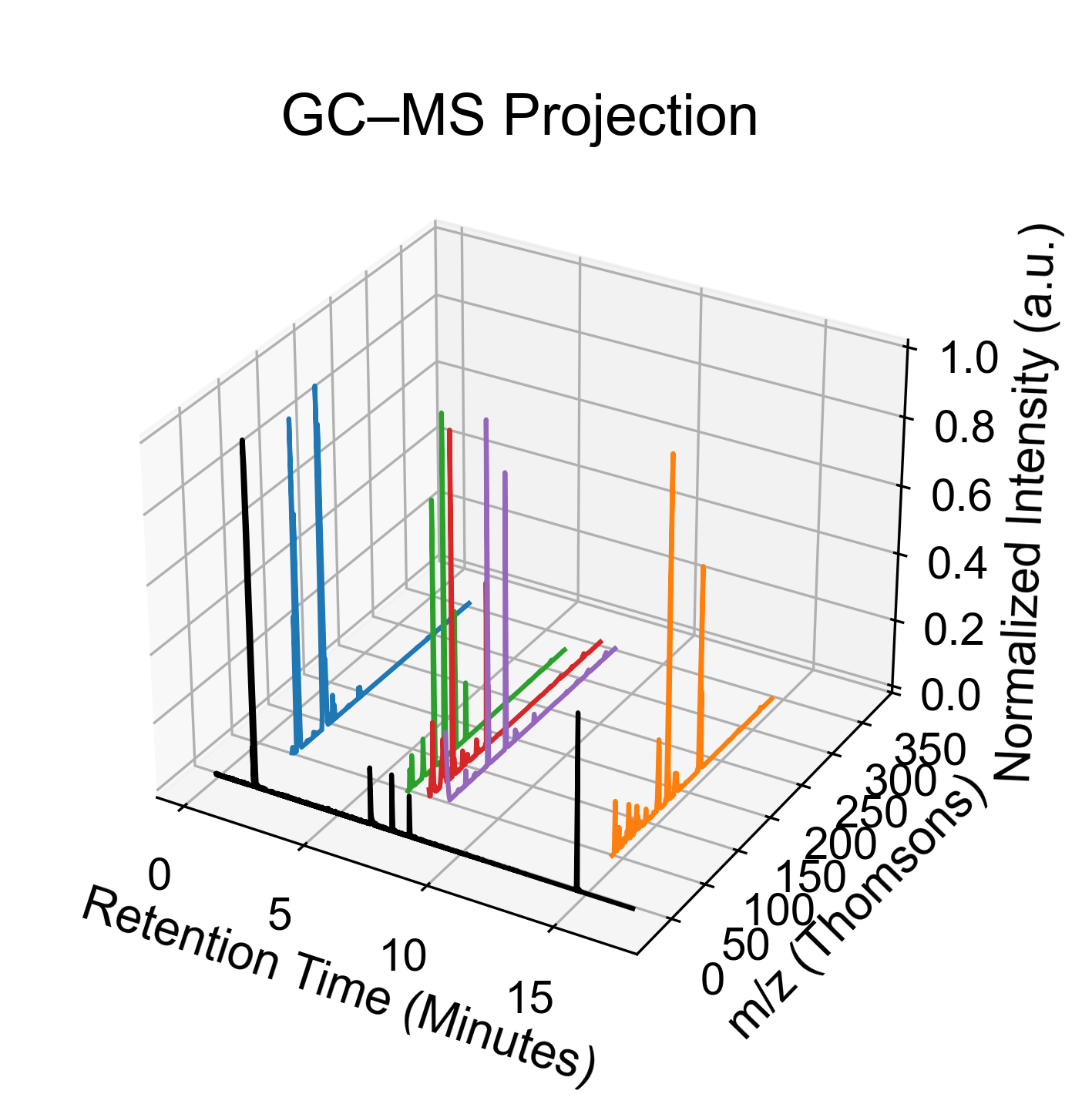}}
\caption{Representative visual comparison of GC-MS spectra between real and generated data. 
(a,b) shows that it preserves the peak position and intensity pattern through the case of \textit{4-nitrophenol + MC}.
(c,d) shows that the proposed peak-aware attention based generative model successfully reconstructs fine-grained peaks even with a complex multi-agent scenario,\textit{2-CEES + 2-CEPS + DFP + DMMP + MC}.}
\label{visual_compare}
\end{figure*}

\subsection{Evaluation metrics}

Quantitative evaluation of generated GC-MS data was conducted using four spectrum-level metrics and two detection-level metrics:

\begin{itemize}
    \item \textbf{Cosine Similarity:} Using cosine values for angles between two vectors in the original GC-MS data and the synthetic GC-MS data, we measure the similarity of the vector~\cite{nakamura2013shape, yoon2024pioneering}.
    
    \item \textbf{Peak Count:} A technique to evaluate peak diversity by comparing characteristic peaks between original GC-MS data and synthetic GC-MS data~\cite{yoon2024pioneering}.
    
    \item \textbf{Pearson Correlation Coefficient (PCC):} Measure the linear strength and orientation-based correlation between the source and the composite GC-MS~\cite{berthold2016clustering, yoon2024pioneering}. 
    
    \item \textbf{3D Visualization Metric:} A custom visualization-based metric implemented in Python \texttt{Matplotlib} and \texttt{Pandas}~\cite{yoon2024pioneering}, which provides a visualized structural comparison  of retention time and $m/z$ distributions~\cite{yoon2024pioneering}.
    
    \item \textbf{Accuracy:} Measure the percentage of correctly detected GC-MS data to the total number of data and use it as a performance indicator for the detection model~\cite{vujovic2021classification}.

    \item \textbf{F1-score:} When class imbalances in the dataset are severe, it is a useful score for measuring performance based on the precision and recall of the detection model~\cite{jeni2013facing}.

\end{itemize}

With these metrics, the quality of the generated data is rigorously evaluated using four metrics (\textbf{Cosine Similarity}, \textbf{Peak Count}, \textbf{PCC}, and \textbf{3D Visualization}), while the performance of the proposed discrimination framework is validated with detection-level metrics such as \textbf{accuracy} and \textbf{F1-score}.

\subsection{Performance Evaluation}

\subsubsection{Measurement of generated data quality}

To evaluate the fidelity of the proposed peak-aware attention conditional generation model, we comprehensively compared real GC-MS data with generated GC-MS data under various conditions. 

As shown in Table~\ref{integrated_quality}, we find that the generative model achieves consistently high global similarity, with the cosine similarity to PCC exceeding 0.94 under all conditions, and preserves the number of peaks well.
In particular, MS data evaluation confirms an almost complete reproduction of the critical diagnostic peak with PCC and cosine values close to 1.0 at key retention times.For example, under \textit{2-CEES + 2-CEPS + EtOH}, the major peak at $t=14.81$ min was reconstructed with PCC=0.99 and Cosine=0.99, while in the challenging case of \textit{2-CEPS + DFP + DMMP + THF}, fine peaks at $t=6.56$ and $t=8.13$ min were faithfully reproduced. 

Representative visual comparisons are presented in Fig.~\ref{visual_compare}. For \textit{4-nitrophenol + MC}, we show that the generated spectra are robust, in close agreement with the actual retention time peak and intensity patterns. On the other hand, a complex multi-agent condition \textit{2-CEES + 2-CEPS + DFP + DMMP + MC} highlights the ability of the model to maintain both global structure and fine-grained $m/z$ peaks, successfully preserving diagnostic features important for detection~\cite{yoon2024pioneering}.

These results confirm that the proposed framework ensures both global consistency and local fidelity, making it a reliable approach for GC-MS data generation under diverse conditions.

\begin{figure}[t!]
\centering \makeatletter\IfFileExists{./figures/result_fin.png}{\includegraphics[width=0.8\textwidth]{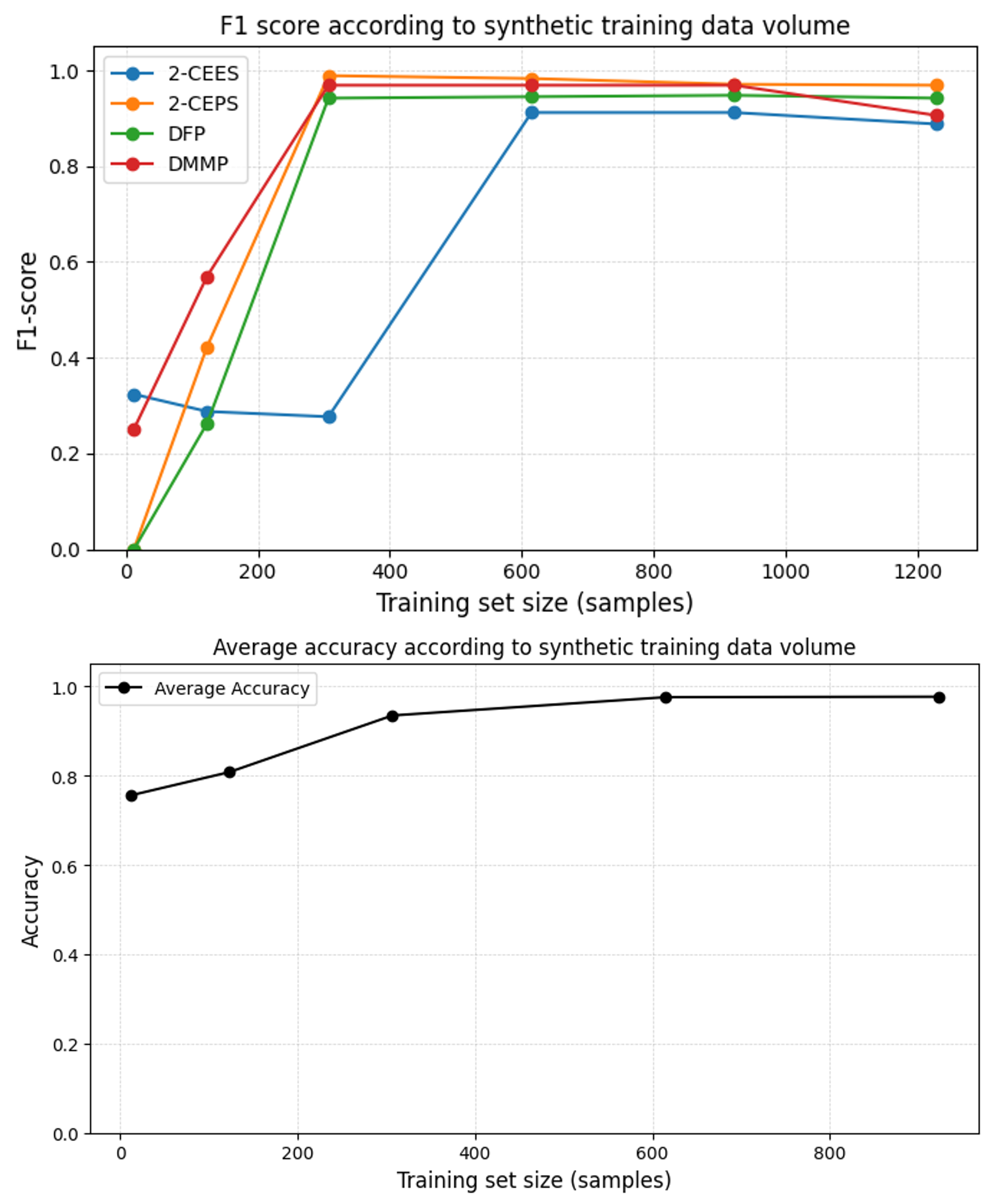}}{}
\makeatother 
\caption{{Changes in detection performance with synthetic training data volume.}}
\label{f1_vs_data}
\end{figure}

\subsubsection{Performance evaluation  of detection framework}

\begin{table*}[ht]
\centering
\caption{The effect of synthetic training data volume on detection performance. The results are precision (P), recall (R), and F1 scores (F1) for each agent, including detection accuracy for the whole. And the average F1 score is the average for all six agents.}
\setlength{\tabcolsep}{3.3pt}
\renewcommand{\arraystretch}{1.12}
\resizebox{\textwidth}{!}{
\begin{tabular}{c|c|
ccc|ccc|ccc|ccc|ccc|ccc|c}
\hline
\multirow{2}{*}{\textbf{Train/Val Size}} & \multirow{2}{*}{\textbf{Accuracy}}
& \multicolumn{3}{c|}{\textbf{2-CEES}}
& \multicolumn{3}{c|}{\textbf{2-CEPS}}
& \multicolumn{3}{c|}{\textbf{DFP}}
& \multicolumn{3}{c|}{\textbf{DMMP}}
& \multicolumn{3}{c|}{\textbf{4-Nitrophenol}}
& \multicolumn{3}{c|}{\textbf{Ethylenediamine}}
& \multirow{2}{*}{\textbf{Avg. F1}} \\
\cline{3-20}
& & \textbf{P} & \textbf{R} & \textbf{F1}
  & \textbf{P} & \textbf{R} & \textbf{F1}
  & \textbf{P} & \textbf{R} & \textbf{F1}
  & \textbf{P} & \textbf{R} & \textbf{F1}
  & \textbf{P} & \textbf{R} & \textbf{F1}
  & \textbf{P} & \textbf{R} & \textbf{F1}
  & \\
\hline
12 / 4     & 0.756 & 0.527 & 0.234 & 0.324 & 0.000 & 0.000 & 0.000 & 0.000 & 0.000 & 0.000 & 0.208 & 0.314 & 0.251 & 0.000 & 0.000 & 0.000 & 0.000 & 0.000 & 0.000 & 0.096 \\
123 / 41   & 0.808 & 0.548 & 0.195 & 0.288 & 0.841 & 0.283 & 0.423 & 0.467 & 0.183 & 0.263 & 0.481 & 0.701 & 0.570 & 0.000 & 0.000 & 0.000 & 1.000 & 0.300 & 0.462 & 0.334 \\
307 / 103  & 0.935 & 0.551 & 0.185 & 0.277 & 0.967 & 0.995 & 0.989 & 1.000 & 0.890 & 0.942 & 0.984 & 0.954 & 0.969 & 0.000 & 0.000 & 0.000 & 1.000 & 0.300 & 0.462 & 0.606 \\
615 / 205  & 0.976 & 0.912 & 0.912 & 0.912 & 0.971 & 0.995 & 0.983 & 0.994 & 0.905 & 0.945 & 0.964 & 0.974 & 0.969 & 0.400 & 0.750 & 0.522 & 0.917 & 0.550 & 0.688 & 0.836 \\
922 / 308  & 0.977 & 0.912 & 0.912 & 0.912 & 0.949 & 0.995 & 0.971 & 0.989 & 0.911 & 0.948 & 0.964 & 0.974 & 0.969 & 0.583 & 0.875 & 0.700 & 1.000 & 0.550 & 0.710 & 0.868 \\

\hline
\end{tabular}}
\label{data_volume_effect}
\end{table*}

We experiment with a detection framework that aims to accurately detect 676 real data points measured in various combinations based on six chemicals.  
Furthermore, to systematically evaluate the proposed detection framework, we have systematically increased the number of synthetic GC-MS data used to train detection models. These steps appear in Table~\ref{data_volume_effect} and, as described in Fig.~\ref{f1_vs_data}, larger synthetic datasets consistently improve detection accuracy and F1-scores across all agents.
Even in the detection framework with state-of-the-art Transformer-based neural networks~\cite{wen2022transformers}, the performance of just 12 training samples was not good, with an average accuracy of F1=0.144 and 0.756, highlighting the limitations of data scarcity. However, after the training set reached 307 samples, the detector achieved an average F1-score of 0.794 and an accuracy of 0.935, showing significant improvements. With over 615 samples, the performance converged to near-optimal levels, demonstrating the reliability and necessity of synthetic data generation models in enhancing model robustness.

Overall, these results confirm the contribution of the proposed peak-aware attention based detection framework, which not only reduces measurement resources by reproducing high-fidelity data but also significantly improves detection robustness under interference conditions.
However, despite 922 synthetic data showing the best performance of blisters and nerve agents, the rather low P results of 4-Nitrophenol and the lack of R results and F1 scores of ethylenediamine suggest a path forward for future work.



\section{Conclusion}

In this work, we presented a framework for conditional generation of peak-aware attention on GC-MS data, which is designed to address the critical challenges of data scarcity and interference complexity in chemical detection. Through a combination of conditional embedding, adversarial learning, and peak-aware attention, the proposed model successfully generated a high-fidelity synthetic spectrum that maintains both global consistency and fine-grained diagnostic peaks. Quantitative results demonstrate cosine similarity and PCC values exceeding 0.94 under various conditions, and faithful reconstruction of key features through visual comparison.

We further validated its usefulness by incorporating the generated data into a transformer-based detection framework. Experiments show that increasing the amount of synthetic data systematically improves the detection performance, and that when the training set exceeds 615 samples, the F1 score and accuracy converge to near-optimal levels. These findings show that the proposed framework makes a dual contribution, reducing measurement requirements and significantly improving the robustness of chemical detection under complex real-world conditions through realistic data generation. Future work will address remaining limitations such as 4-nitrophenol and ethylenediamine, paving the way to a more universally reliable detection system.

\section*{Acknowledgments}
This work was supported by Korea Research Institute of defense Technology planning and advancement (KRIT) grant funded by Defense Acquisition Program Administration (DAPA) (KRIT-CT-21-034).

\bibliographystyle{cas-model2-names}

\bibliography{cas-ref}

@article{valdez2018analysis,
  title={Analysis of chemical warfare agents by gas chromatography-mass spectrometry: methods for their direct detection and derivatization approaches for the analysis of their degradation products},
  author={Valdez, Carlos A and Leif, Roald N and Hok, Saphon and Hart, Bradley R},
  journal={Reviews in Analytical Chemistry},
  volume={37},
  number={1},
  pages={20170007},
  year={2018},
  publisher={De Gruyter}
}

@article{mavstovska2003practical,
  title={Practical approaches to fast gas chromatography--mass spectrometry},
  author={Ma{\v{s}}tovsk{\'a}, Kate{\v{r}}ina and Lehotay, Steven J},
  journal={Journal of Chromatography A},
  volume={1000},
  number={1-2},
  pages={153--180},
  year={2003},
  publisher={Elsevier}
}

@article{ranjan2023gas,
  title={Gas chromatography--mass spectrometry (GC-MS): A comprehensive review of synergistic combinations and their applications in the past two decades},
  author={RANJAN, Smriti and Chaitali, ROY and SINHA, Sandip KUMAR},
  journal={Journal of Analytical Sciences and Applied Biotechnology},
  volume={5},
  number={2},
  pages={72--85},
  year={2023}
}

@inproceedings{yoon2024pioneering,
  title={Pioneering ai in chemical data: New frontline with gc-ms generation},
  author={Yoon, Namkyung and Kim, Hwangnam},
  booktitle={2024 International Conference on Artificial Intelligence in Information and Communication (ICAIIC)},
  pages={826--831},
  year={2024},
  organization={IEEE}
}

@article{yoon2024unveiling,
  title={Unveiling hidden insights in gas chromatography data analysis with generative adversarial networks},
  author={Yoon, Namkyung and Jung, Wooyong and Kim, Hwangnam},
  journal={Chemosensors},
  volume={12},
  number={7},
  pages={131},
  year={2024},
  publisher={MDPI}
}

@article{berahmand2024autoencoders,
  title={Autoencoders and their applications in machine learning: a survey},
  author={Berahmand, Kamal and Daneshfar, Fatemeh and Salehi, Elaheh Sadat and Li, Yuefeng and Xu, Yue},
  journal={Artificial intelligence review},
  volume={57},
  number={2},
  pages={28},
  year={2024},
  publisher={Springer}
}

@article{cordonnier2020multi,
  title={Multi-head attention: Collaborate instead of concatenate},
  author={Cordonnier, Jean-Baptiste and Loukas, Andreas and Jaggi, Martin},
  journal={arXiv preprint arXiv:2006.16362},
  year={2020}
}

@article{durak2003short,
  title={Short-time Fourier transform: two fundamental properties and an optimal implementation},
  author={Durak, Lutfiye and Arikan, Orhan},
  journal={IEEE Transactions on Signal Processing},
  volume={51},
  number={5},
  pages={1231--1242},
  year={2003},
  publisher={IEEE}
}

@article{haber2017simple,
  title={A simple multi-objective optimization based on the cross-entropy method},
  author={Haber, Rodolfo E and Beruvides, Gerardo and Quiza, Ram{\'o}n and Hernandez, Alejandro},
  journal={Ieee Access},
  volume={5},
  pages={22272--22281},
  year={2017},
  publisher={IEEE}
}

@article{smith2020conditional,
  title={Conditional GAN for timeseries generation},
  author={Smith, Kaleb E and Smith, Anthony O},
  journal={arXiv preprint arXiv:2006.16477},
  year={2020}
}

@article{hodson2022root,
  title={Root mean square error (RMSE) or mean absolute error (MAE): When to use them or not},
  author={Hodson, Timothy O},
  journal={Geoscientific Model Development Discussions},
  volume={2022},
  pages={1--10},
  year={2022},
  publisher={G{\"o}ttingen, Germany}
}

@inproceedings{yoon2024dmgan,
  title={DMGAN: Bridging AI and Chemistry with Enhanced GC-MS Data Generation},
  author={Yoon, Namkyung and Kim, Hwangnam},
  booktitle={2024 IEEE SENSORS},
  pages={1--4},
  year={2024},
  organization={IEEE}
}

@article{yoon2024detecting,
  title={Detecting DDoS based on attention mechanism for Software-Defined Networks},
  author={Yoon, Namkyung and Kim, Hwangnam},
  journal={Journal of Network and Computer Applications},
  volume={230},
  pages={103928},
  year={2024},
  publisher={Elsevier}
}

@article{yoon2024adaptive,
  title={Adaptive sensing data augmentation for drones using attention-based gan},
  author={Yoon, Namkyung and Kim, Kiseok and Lee, Sangmin and Bai, Jin Hyoung and Kim, Hwangnam},
  journal={Sensors},
  volume={24},
  number={16},
  pages={5451},
  year={2024},
  publisher={MDPI}
}

@article{yoon2023steam,
  title={STEAM: Spatial Trajectory Enhanced Attention Mechanism for Abnormal UAV Trajectory Detection},
  author={Yoon, Namkyung and Lee, Dongjae and Kim, Kiseok and Yoo, Taehoon and Joo, Hyeontae and Kim, Hwangnam},
  journal={Applied Sciences},
  volume={14},
  number={1},
  pages={248},
  year={2023},
  publisher={MDPI}
}

@article{yoon2025relaygan,
  title={RelayGAN: Sequential knowledge propagation for sustainable multi-generation},
  author={Yoon, Namkyung and Kim, Hwangnam},
  journal={Array},
  pages={100444},
  year={2025},
  publisher={Elsevier}
}

@article{singh2021overview,
  title={An overview of variational autoencoders for source separation, finance, and bio-signal applications},
  author={Singh, Aman and Ogunfunmi, Tokunbo},
  journal={Entropy},
  volume={24},
  number={1},
  pages={55},
  year={2021},
  publisher={MDPI}
}

@inproceedings{audebert2018generative,
  title={Generative adversarial networks for realistic synthesis of hyperspectral samples},
  author={Audebert, Nicolas and Le Saux, Bertrand and Lef{\`e}vre, S{\'e}bastien},
  booktitle={IGARSS 2018-2018 IEEE International Geoscience and Remote Sensing Symposium},
  pages={4359--4362},
  year={2018},
  organization={IEEE}
}

@article{vig2019analyzing,
  title={Analyzing the structure of attention in a transformer language model},
  author={Vig, Jesse and Belinkov, Yonatan},
  journal={arXiv preprint arXiv:1906.04284},
  year={2019}
}

@article{brauwers2021general,
  title={A general survey on attention mechanisms in deep learning},
  author={Brauwers, Gianni and Frasincar, Flavius},
  journal={IEEE transactions on knowledge and data engineering},
  volume={35},
  number={4},
  pages={3279--3298},
  year={2021},
  publisher={IEEE}
}

@article{mirza2014conditional,
  title={Conditional generative adversarial nets},
  author={Mirza, Mehdi and Osindero, Simon},
  journal={arXiv preprint arXiv:1411.1784},
  year={2014}
}

@article{zheng2010advances,
  title={Advances in the chemical sensors for the detection of DMMP—A simulant for nerve agent sarin},
  author={Zheng, Qi and Fu, Yong-chun and Xu, Jia-qiang},
  journal={Procedia Engineering},
  volume={7},
  pages={179--184},
  year={2010},
  publisher={Elsevier}
}

@article{wymore2014hydrolysis,
  title={Hydrolysis of DFP and the nerve agent (S)-sarin by DFPase proceeds along two different reaction pathways: implications for engineering bioscavengers},
  author={Wymore, Troy and Field, Martin J and Langan, Paul and Smith, Jeremy C and Parks, Jerry M},
  journal={The Journal of Physical Chemistry B},
  volume={118},
  number={17},
  pages={4479--4489},
  year={2014},
  publisher={ACS Publications}
}

@article{kim2021decomposition,
  title={Decomposition of the simulant 2-chloroethyl ethyl sulfide blister agent under ambient conditions using metal--organic frameworks},
  author={Kim, Hong-Hyun and Seo, Jin Young and Kim, Heejeong and Jeong, Sangjo and Baek, Kyung-Youl and Kim, Jongsik and Min, Sein and Kim, Sang Hoon and Jeong, Keunhong},
  journal={ACS applied materials \& interfaces},
  volume={13},
  number={3},
  pages={3782--3792},
  year={2021},
  publisher={ACS Publications}
}

@article{dubey2009thiocompounds,
  title={Thiocompounds as simulants of sulphur mustard for testing of protective barriers},
  author={Dubey, Vinita and Parmar, Toral and Saxena, Amit and Agarwal, DD},
  journal={Journal of Applied Polymer Science},
  volume={111},
  number={2},
  pages={928--933},
  year={2009},
  publisher={Wiley Online Library}
}

@article{apak2020colorimetric,
  title={Colorimetric sensors and nanoprobes for characterizing antioxidant and energetic substances},
  author={Apak, Re{\c{s}}at and {\c{C}}eki{\c{c}}, Sema Demirci and {\"U}zer, Ay{\c{s}}em and {\c{C}}apano{\u{g}}lu, Esra and {\c{C}}elik, Saliha Esin and Bener, Mustafa and Can, Ziya and Durmazel, Selen},
  journal={Analytical Methods},
  volume={12},
  number={44},
  pages={5266--5321},
  year={2020},
  publisher={Royal Society of Chemistry}
}

@article{choodum2017selective,
  title={Selective pre and post blast trinitrotoluene detection with a novel ethylenediamine entrapped thin polymer film and digital image colorimetry},
  author={Choodum, Aree and Keson, Jutaporn and Kanatharana, Proespichaya and Limsakul, Wadcharawadee and Wongniramaikul, Worawit},
  journal={Sensors and Actuators B: Chemical},
  volume={252},
  pages={463--469},
  year={2017},
  publisher={Elsevier}
}

@inproceedings{jeni2013facing,
  title={Facing imbalanced data--recommendations for the use of performance metrics},
  author={Jeni, L{\'a}szl{\'o} A and Cohn, Jeffrey F and De La Torre, Fernando},
  booktitle={2013 Humaine association conference on affective computing and intelligent interaction},
  pages={245--251},
  year={2013},
  organization={IEEE}
}

@article{wen2022transformers,
  title={Transformers in time series: A survey},
  author={Wen, Qingsong and Zhou, Tian and Zhang, Chaoli and Chen, Weiqi and Ma, Ziqing and Yan, Junchi and Sun, Liang},
  journal={arXiv preprint arXiv:2202.07125},
  year={2022}
}

@article{vujovic2021classification,
  title={Classification model evaluation metrics},
  author={Vujovi{\'c}, {\v{Z}}eljko and others},
  journal={International Journal of Advanced Computer Science and Applications},
  volume={12},
  number={6},
  pages={599--606},
  year={2021}
}

@article{berthold2016clustering,
  title={On clustering time series using euclidean distance and pearson correlation},
  author={Berthold, Michael R and H{\"o}ppner, Frank},
  journal={arXiv preprint arXiv:1601.02213},
  year={2016}
}

@article{nakamura2013shape,
  title={A shape-based similarity measure for time series data with ensemble learning},
  author={Nakamura, Tetsuya and Taki, Keishi and Nomiya, Hiroki and Seki, Kazuhiro and Uehara, Kuniaki},
  journal={Pattern Analysis and Applications},
  volume={16},
  number={4},
  pages={535--548},
  year={2013},
  publisher={Springer}
}

@article{eckenrode2001environmental,
  title={Environmental and forensic applications of field-portable GC-MS: an overview},
  author={Eckenrode, BA},
  journal={Journal of the American Society for Mass Spectrometry},
  volume={12},
  number={6},
  pages={683--693},
  year={2001},
  publisher={ACS Publications}
}

@article{yoon2019time,
  title={Time-series generative adversarial networks},
  author={Yoon, Jinsung and Jarrett, Daniel and Van der Schaar, Mihaela},
  journal={Advances in neural information processing systems},
  volume={32},
  year={2019}
}

@article{zhu2019electrocardiogram,
  title={Electrocardiogram generation with a bidirectional LSTM-CNN generative adversarial network},
  author={Zhu, Fei and Ye, Fei and Fu, Yuchen and Liu, Quan and Shen, Bairong},
  journal={Scientific reports},
  volume={9},
  number={1},
  pages={6734},
  year={2019},
  publisher={Nature Publishing Group UK London}
}

@article{dewi2022synthetic,
  title={Synthetic Data generation using DCGAN for improved traffic sign recognition},
  author={Dewi, Christine and Chen, Rung-Ching and Liu, Yan-Ting and Tai, Shao-Kuo},
  journal={Neural Computing and Applications},
  volume={34},
  number={24},
  pages={21465--21480},
  year={2022},
  publisher={Springer}
}

@article{mclafferty1978separation,
  title={Separation/identification system for complex mixtures using mass separation and mass spectral characterization},
  author={McLafferty, Fred W and Bockhoff, Frank M},
  journal={Analytical Chemistry},
  volume={50},
  number={1},
  pages={69--76},
  year={1978},
  publisher={ACS Publications}
}

@book{hubschmann2025handbook,
  title={Handbook of GC-MS: fundamentals and applications},
  author={H{\"u}bschmann, Hans-Joachim},
  year={2025},
  publisher={John Wiley \& Sons}
}

@article{camarasu2000headspace,
  title={Headspace SPME method development for the analysis of volatile polar residual solvents by GC-MS},
  author={Camarasu, Costin C},
  journal={Journal of pharmaceutical and biomedical analysis},
  volume={23},
  number={1},
  pages={197--210},
  year={2000},
  publisher={Elsevier}
}

@article{quan2024deep,
  title={Deep learning-based image and video inpainting: A survey},
  author={Quan, Weize and Chen, Jiaxi and Liu, Yanli and Yan, Dong-Ming and Wonka, Peter},
  journal={International Journal of Computer Vision},
  volume={132},
  number={7},
  pages={2367--2400},
  year={2024},
  publisher={Springer}
}


\end{document}